\documentstyle[aclap]{article}

\makeatletter
\def\section{\@startsection {section}{1}{\z@}
    {-2ex plus -.3ex minus -.1ex}{1ex plus .1ex}{\Large\bf}}
\def\subsection{\@startsection{subsection}{2}{\z@}
    {-1.5ex plus -.4ex minus -.1ex}{1ex plus .1ex}{\large\bf}}
\long\def\@makecaption#1#2{
 \vskip .2ex 
 \setbox\@tempboxa\hbox{#1: #2}
 \ifdim \wd\@tempboxa >\hsize #1: #2\par \else \hbox
to\hsize{\hfil\box\@tempboxa\hfil} 
 \fi}
\makeatother

\newcommand{\pos}{{\sc pos}}
\newcommand{\ignore}[1]{}
\newcommand{\ra}{$\rightarrow$}

\newcommand{\andrei}{\scriptsize\bf \setlength{\parindent}{0mm} }
\newcommand{\longline}{\rule{\textwidth}{0.01in}}

\title{\vspace{-0.5in}
  Unsupervised Learning of Word-Category Guessing Rules}

\author{Andrei Mikheev  \\
\\  HCRC Language Technology Group \\ University of Edinburgh \\
2 Buccleuch Place \\ Edinburgh EH8 9LW, Scotland, UK \\ 
{\tt: Andrei.Mikheev@ed.ac.uk}}

\begin{document}
\bibliographystyle{fullname}
\maketitle
\vspace{-0.5in}
\begin{abstract}
  Words unknown to the lexicon present a substantial problem to
  part-of-speech tagging.  In this paper we present a technique for fully
  unsupervised statistical acquisition of rules which guess possible
  parts-of-speech for unknown words.  Three complementary sets of
  word-guessing rules are induced from the lexicon and a raw corpus:
  prefix morphological rules, suffix morphological rules and
  ending-guessing rules. The learning was performed on the Brown Corpus
  data and rule-sets, with a highly competitive performance, were
  produced and compared with the state-of-the-art.
\end{abstract}

\section{Introduction}
Words unknown to the lexicon present a substantial problem to
part-of-speech (\pos ) tagging of real-world texts.  Taggers assign a
single \pos -tag to a word-token, provided that it is known what
parts-of-speech this word can take on in principle. So, first words are
looked up in the lexicon.  However, 3 to 5\% of word tokens are usually
missing in the lexicon when tagging real-world texts.  This is where
word-\pos\ guessers take their place --- they employ the analysis of word
features, e.g.  word leading and trailing characters, to figure out its
possible \pos\ categories.  A set of rules which on the basis of ending
characters of unknown words, assign them with sets of possible \pos -tags
is supplied with the Xerox tagger \cite{Kupiec:1992}.  A similar approach
was taken in \cite{Weischedel:1993} where an unknown word was guessed
given the probabilities for an unknown word to be of a particular \pos ,
its capitalisation feature and its ending.  In \cite{Brill:1995} a system
of rules which uses both ending-guessing and more morphologically
motivated rules is described.  The best of these methods are reported to
achieve 82--85\% of tagging accuracy on unknown words, e.g.
\cite{Brill:1995,Weischedel:1993}.

The major topic in the development of word-\pos\ guessers is the strategy
which is to be used for the acquisition of the guessing rules.  A
rule-based tagger described in \cite{Voutilainen:1995} is equipped with a
set of guessing rules which has been hand-crafted using knowledge of
English morphology and intuition.  A more appealing approach is an
empirical automatic acquisition of such rules using available lexical
resources.  In \cite{Zhang:1990} a system for the automated learning of
morphological word-formation rules is described. This system divides a
string into three regions and from training examples infers their
correspondence to underlying morphological features.  Brill
\cite{Brill:1995} outlines a transformation-based learner which learns
guessing rules from a pre-tagged training corpus.  A statistical-based
suffix learner is presented in \cite{Schmid:1994}. From a pre-tagged
training corpus it constructs the suffix tree where every suffix is
associated with its information measure.  Although the learning process
in these and some other systems is fully unsupervised and the accuracy of
obtained rules reaches current state-of-the-art, they require specially
prepared training data --- a pre-tagged training corpus, training
examples, etc.

In this paper we describe a new fully automatic technique for learning
part-of-speech guessing rules. This technique does not require specially
prepared training data and employs fully unsupervised statistical
learning using the lexicon supplied with the tagger and word-frequencies
obtained from a raw corpus. The learning is implemented as a two-staged
process with feedback. First, setting certain parameters a set of
guessing rules is acquired, then it is evaluated and the results of
evaluation are used for re-acquisition of a better tuned rule-set.

\section{Guessing Rules Acquisition}
 
As was pointed out above, one of the requirements in many techniques for
automatic learning of part-of-speech guessing rules is specially prepared
training data --- a pre-tagged training corpus, training examples, etc.
In our approach we decided to reuse the data which come naturally with a
tagger, viz. the lexicon.  Another source of information which is used
and which is not prepared specially for the task is a text corpus. Unlike
other approaches we don't require the corpus to be pre-annotated but use
it in its raw form.  In our experiments we used the lexicon and
word-frequencies derived from the Brown Corpus \cite{Francis:1982}.
There are a number of reasons for choosing the Brown Corpus data for
training.  The most important ones are that the Brown Corpus provides a
model of general multi-domain language use, so general language
regularities can be induced from it, and second, many taggers come with
data trained on the Brown Corpus which is useful for comparison and
evaluation. This, however, by no means restricts the described technique
to that or any other tag-set, lexicon or corpus.  Moreover, despite the
fact that the training is performed on a particular lexicon and a
particular corpus, the obtained guessing rules suppose to be domain and
corpus independent and the only training-dependent feature is the tag-set
in use.

The acquisition of word-\pos\ guessing rules is a three-step procedure
which includes the {\em rule extraction}, {\em rule scoring} and {\em
  rule merging} phases.  At the rule extraction phase, three sets of
word-guessing rules (morphological prefix guessing rules, morphological
suffix guessing rules and ending-guessing rules) are extracted from the
lexicon and cleaned from coincidental cases.  At the scoring phase, each
rule is scored in accordance with its accuracy of guessing and the best
scored rules are included into the final rule-sets.  At the merging
phase, rules which have not scored high enough to be included into the
final rule-sets are merged into more general rules, then re-scored and
depending on their score added to the final rule-sets.

\subsection{Rule Extraction Phase}

\subsubsection {Extraction of Morphological  Rules.}
\label{sec:morph}

Morphological word-guessing rules describe how one word can be guessed
given that another word is known. For example, the rule: {\andrei [un
  (VBD VBN) (JJ)] } says that prefixing the string ``un'' to a word,
which can act as past form of verb (VBD) and participle (VBN), produces
an adjective (JJ).  For instance, by applying this rule to the word
``undeveloped'', we first segment the prefix ``un'' and if the remaining
part ``developed'' is found in the lexicon as {\andrei (VBD VBN)}, we
conclude that the word ``undeveloped'' is an adjective (JJ).  The first
\pos -set in a guessing rule is called the {\em initial class}
($I$-class) and the \pos -set of the guessed word is called the {\em
  resulting class} ($R$-class).  In the example above {\andrei (VBD VBN)}
is the $I$-class of the rule and {\andrei (JJ)} is the $R$-class.

In English, as in many other languages, morphological word formation is
realised by affixation: prefixation and suffixation.  Although sometimes
the affixation is not just a straightforward concatenation of the affix
with the stem\footnote{consider an example: try -- tried.}, the majority
of cases clearly obey simple concatenative regularities.  So, we decided
first to concentrate only on simple concatenative cases.  There are two
kinds of morphological rules to be learned: suffix rules ($A^{s}$) ---
rules which are applied to the tail of a word, and prefix rules ($A^{p}$)
--- rules which are applied to the beginning of a word.  For example:

$A^{s}$ :  {\andrei [ed (NN VB) (JJ VBD VBN)]}

says that if by stripping the suffix ``ed'' from an unknown word we
produce a word with the \pos -class {\andrei (NN VB)}, the unknown word
is of the class -- {\andrei (JJ VBD VBN)}.  This rule works, for
instance, for {\andrei [book \ra booked], [water \ra watered]}, etc.  To
extract such rules a special operator $\bigtriangledown$ is applied to
every pair of words from the lexicon.  It tries to segment an affix by
leftmost string subtraction for suffixes and rightmost string subtraction
for prefixes. If the subtraction results in an non-empty string it
creates a morphological rule by storing the \pos -class of the shorter
word as the $I$-class and the \pos -class of the longer word as the
$R$-class. For example:

 
{\andrei [booked (JJ VBD VBN)]} $\bigtriangledown$
{\andrei [book (NN VB)]}  \ra  \\
\hspace*{18ex}  $A^{s}:${\andrei [ed (NN VB) (JJ VBD VBN)]} 

{\andrei [undeveloped (JJ)]} $\bigtriangledown$ 
{\andrei [developed (VBD VBN)]}   \ra \\ 
\hspace*{25ex} $A^{p}:${\andrei [un (VBD VBN) (JJ)]}


The $\bigtriangledown$ operator is applied to all possible lexicon-entry
pairs and if a rule produced by such an application has already been
extracted from another pair, its frequency count ($f$) is incremented.
Thus two different sets of guessing rules --- prefix and suffix
morphological rules together with their frequencies --- are produced.
Next, from these sets of guessing rules we need to cut out infrequent
rules which might bias the further learning process.  To do that we
eliminate all the rules with the frequency $f$ less than a certain
threshold $\theta$\footnote{usually we set this threshold quite low:
  2--4.}.  Such filtering reduces the rule-sets more than tenfold and
does not leave clearly coincidental cases among the rules.

\subsubsection{Extraction of Ending Guessing Rules.}

Unlike morphological guessing rules, ending-guessing rules do not require
the main form of an unknown word to be listed in the lexicon.  These
rules guess a \pos -class for a word just on the basis of its ending
characters and without looking up its stem in the lexicon.  Such rules
are able to cover more unknown words than morphological guessing rules
but their accuracy will not be as high.  For example, an ending-guessing
rule

$A^{e}$:{\andrei  [ing -  (JJ NN VBG)]}

says that if a word ends with ``ing'' it can be an adjective, a noun or a
gerund. Unlike a morphological rule, this rule does not ask to check
whether the substring preceeding the ``ing''-ending is a word with a
particular \pos -tag.  Thus an ending-guessing rule looks exactly like a
morphological rule apart from the $I$-class which is always void.

To collect such rules we set the upper limit on the ending length equal
to five characters and thus collect from the lexicon all possible
word-endings of length 1, 2, 3, 4 and 5, together with the \pos -classes
of the words where these endings were detected to appear. This is done by
the operator $\bigtriangleup$.  For example, from the word {\andrei
  [different (JJ)]} the $\bigtriangleup$ operator will produce five
ending-guessing rules: 
{\andrei [t - (JJ)]; [nt - (JJ)]; [ent - (JJ)]; [rent - (JJ)]; 
[erent - (JJ)].} 
The $\bigtriangleup$ operator is
applied to each entry in the lexicon in the way described for the
$\bigtriangledown$ operator of the morphological rules and then
infrequent rules with $f<\theta$ are filtered out.

\subsection{Rule Scoring Phase}
\label{sec:score}

Of course, not all acquired rules are equally good as plausible guesses
about word-classes: some rules are more accurate in their guessings and
some rules are more frequent in their application.  So, for every
acquired rule we need to estimate whether it is an effective rule which
is worth retaining in the final rule-set.  For such estimation we perform
a statistical experiment as follows: for every rule we calculate the
number of times this rule was applied to a word token from a raw corpus
and the number of times it gave the right answer. Note that the task of
the rule is not to disambiguate a word's \pos\ but to provide {\em all}
and {\em only} possible \pos s it can take on. If the rule is correct in
the majority of times it was applied it is obviously a good rule. If the
rule is wrong most of the times it is a bad rule which should not be
included into the final rule-set.

To perform this experiment we take one-by-one each rule from the
rule-sets produced at the rule extraction phase, take each word token
from the corpus and guess its \pos -set using the rule if the rule is
applicable to the word. For example, if a guessing rule strips a
particular suffix and a current word from the corpus does not have this
suffix, we classify these word and rule as incompatible and the rule as
not applicable to that word. If the rule is applicable to the word we
perform look-up in the lexicon for this word and then compare the result
of the guess with the information listed in the lexicon. If the guessed
\pos -set is the same as the \pos -set stated in the lexicon, we count it
as success, otherwise it is failure.  The value of a guessing rule, thus,
closely correlates with its {\em estimated proportion of success}
($\hat{p}$) which is the proportion of all positive outcomes ($x$) of the
rule application to the total number of the trials ($n$), which are, in
fact, attempts to apply the rule to all the compatible words in the
corpus.  We also smooth $\hat{p}$ so as not to have zeros in positive or
negative outcome probabilities: $ \hat{p} = \frac{x+0.5 }{n+1 } $

$\hat{p}$ estimate is a good indicator of rule accuracy. However, it
frequently suffers from large {\em estimation error} due to insufficient
training data. For example, if a rule was detected to work just twice and
the total number of observations was also two, its estimate $\hat{p}$ is
very high (1, or 0.83 for the smoothed version) but clearly this is not a
very reliable estimate because of the tiny size of the sample.  Several
smoothing methods have been proposed to reduce the estimation error.  For
different reasons all these smoothing methods are not very suitable in
our case. In our approach we tackle this problem by calculating the {\em
  lower confidence limit} $\pi_{L}$ for the rule estimate. This can be
seen as the minimal expected value of $\hat{p}$ for the rule if we were
to draw a large number of samples. Thus with certain confidence $\alpha$
we can assume that if we used more training data, the rule estimate
$\hat{p}$ would be no worse than the $\pi_{L}$ limit. The lower
confidence limit $\pi_{L}$ is calculated as:

$\pi_{L}=\hat{p} - z_{(1-\alpha)/2}* s_{p}= 
\hat{p }- z_{(1-\alpha)/2}*\sqrt{\frac{\hat{p}(1-\hat{p})}{n}} $

This function favours the rules with higher estimates obtained over
larger samples. Even if one rule has a high estimate but that estimate
was obtained over a small sample, another rule with a lower estimate but
over a large sample might be valued higher. Note also that since
$\hat{p}$ itself is smoothed we will not have zeros in positive
($\hat{p}$) or negative ($1-\hat{p}$) outcome probabilities.  This
estimation of the rule value in fact resembles that used by
\cite{Tzoukermann:1995} for scoring \pos -disambiguation rules for the
French tagger. The main difference between the two functions is that
there the $z$ value was implicitly assumed to be 1 which corresponds to
the confidence of 68\%. A more standard approach is to adopt a rather
high confidence value in the range of 90-95\%. We adopted 90\% confidence
for which $z_{(1-0.90)/2} = z_{0.05} = 1.65$. Thus we can calculate the
score for the i{\em th} rule as: 
$ \hat{p}_{i} - 1.65*\sqrt{\frac{\hat{p}_{i}(1-\hat{p}_{i})}{n_{i}}}$

Another important consideration for scoring a word-guessing rule is that
the longer the affix or ending of the rule the more confident we are that
it is not a coincidental one, even on small samples. For example, if the
estimate for the word-ending ``o'' was obtained over a sample of 5 words
and the estimate for the word-ending ``fulness'' was also obtained over a
sample of 5 words, the later case is more representative even though the
sample size is the same. Thus we need to adjust the estimation error in
accordance with the length of the affix or ending.  A good way to do that
is to divide it by a value which increases along with the increase of the
length. After several experiments we obtained:

$score_{i}= \hat{p}_{i} -
1.65*\sqrt{\frac{\hat{p}_{i}(1-\hat{p}_{i})}{n_{i}}}/(1+\log(|S_{i}|))$

When the length of the affix or ending is 1 the estimation error is not
changed since $\log(1)$ is $0$.  For the rules with the affix or ending
length of 2 the estimation error is reduced by $1+\log(2)= 1.3$, for the
length 3 this will be $1+\log(3)=1.48$, etc. The longer the length the
smaller the sample which will be considered representative enough for a
confident rule estimation.  Setting the threshold $\theta_{s}$ at a
certain level lets only the rules whose score is higher than the
threshold to be included into the final rule-sets.  The method for
setting up this threshold is based on empirical evaluations of the
rule-sets and is described in Section~\ref{sec:eval}.

\subsection{Rule Merging Phase}

Rules which have scored lower than the threshold $\theta_{s}$ can be
merged into more general rules which if scored above the threshold are
also included into the final rule-sets. We can merge two rules which have
scored below the threshold and have the same affix (or ending) and the
initial class ({\em I})\footnote{For ending-guessing rules this is always
  true, so only the ending itself counts.}.  The score of the resulting
rule will be higher than the scores of the merged rules since the number
of positive observations increases and the number of the trials remains
the same.  After a successful application of the merging, the resulting
rule substitutes the two merged ones. To perform such rule-merging over a
rule-set, first, the rules which have not been included into the final
set are sorted by their score and best-scored rules are merged first.
This is done recursively until the score of the resulting rule does not
exceed the threshold in which case it is added to the final rule-set.
This process is applied until no merges can be done to the rules which
have scored below the threshold.

\section{Direct Evaluation Stage}
\label{sec:eval}

\begin{table*}[h]  
\begin{minipage}{\hsize}
\begin{tabular}[t]{||l|l|l|l|l|l||}
\hline\hline
 Measure     & Test      & Xerox     &  Ending 75& Suffix 60 & Prefix 80\\ 
\hline
 Recall       & Lexicon  & 0.956313  &  0.945726 & 0.95761   & 0.955748 \\
              & Corpus   & 0.944526  &  0.952016 & 0.97352   & 0.978515 \\
\hline
 Precision    & Lexicon  & 0.460761  &  0.675122 &0.919796   & 0.922534 \\
              & Corpus   & 0.523965  &  0.745339 & 0.979351  & 0.977633 \\
\hline
Coverage      & Lexicon  & 0.917698  &  0.977089 & 0.37597   & 0.049558 \\
              & Corpus   & 0.893275  &  0.96104  & 0.320996  & 0.058372  \\
\hline\hline
\end{tabular}

\vspace{2ex} 

\label{tab:eval_sure_lex}

\caption{Results obtained at the evaluation of
  the acquired rule-sets over the training lexicon and the training
  corpus.  Guessing rule-sets produced using different confidence
  thresholds were compared. Best-scored rule-sets detected: Prefix 80 -
  prefix morphological rules which scored over 80 points, Suffix 60 -
  suffix morphological rules which scored over 60 points and Ending 75 -
  ending-guessing rules which scored over 75 points.  As the base-line
  model was taken the ending guesser developed by Xerox (X).}

\longline
\end{minipage}
\end{table*}

There are two important questions which arise at the rule acquisition
stage - how to choose the scoring threshold $\theta_{s}$ and what is the
performance of the rule-sets produced with different thresholds.  The
task of assigning a set of \pos -tags to a word is actually quite similar
to the task of document categorisation where a document should be
assigned with a set of descriptors which represent its contents.  The
performance of such assignment can be measured in:
 
{\em recall} - the percentage of \pos s which were assigned
  correctly by the guesser to a word;

{\em precision} - the percentage of \pos s the  guesser
  assigned correctly  over the total number of \pos s
  it assigned to the word;

{\em coverage} - the proportion of words which the guesser was
able to classify, but not necessarily  correctly;


In our experiments we measured word precision and word recall
(micro-average).  There were two types of data in use at this stage.
First, we evaluated the guessing rules against the actual lexicon: every
word from the lexicon, except for closed-class words and words shorter
than five characters\footnote{the actual size of the filtered lexicon was
  47,659 entries out of 53,015 entries of the original lexicon.}, was
guessed by the different guessing strategies and the results were
compared with the information the word had in the lexicon.  In the other
evaluation experiment we measured the performance of the guessing rules
against the training corpus.  For every word we computed its metrics
exactly as in the previous experiment. Then we multiplied these results
by the corpus frequency of this particular word and averaged them. Thus
the most frequent words had the greatest influence on the aggreagte
measures.

First, we concentrated on finding the best thresholds $\theta_{s}$ for
the rule-sets. To do that for each rule-set produced using different
thresholds we recorded the three metrics and chose the set with the best
aggregate.  In \ignore{ Table~\ref{tab:eval_sure_lex} } Table~1 some
results of that experiment are shown.  The best thresholds were detected:
for ending rules -- 75 points, for suffix rules -- 60, and for prefix
rules -- 80.  One can notice a slight difference in the results obtained
over the lexicon and the corpus. The corpus results are better because
the training technique explicitly targeted the rule-sets to the most
frequent cases of the corpus rather than the lexicon.  In average
ending-guessing rules were detected to cover over 96\% of the unknown
words. The precision of 74\% roughly can be interpreted as that for words
which take on three different \pos s in their \pos -class, the
ending-guessing rules will assign four, but in 95\% of the times (recall)
the three required \pos s will be among the four assigned by the guess.
In comparison with the Xerox word-ending guesser taken as the base-line
model we detect a substantial increase in the precision by about 22\% and
a cheerful increase in coverage by about 6\%. This means that the Xerox
guesser creates more ambiguity for the disambiguator, assigning five
instead of three \pos s in the example above. It can also handle 6\% less
unknown words which, in fact, might decrease its performance even lower.
In comparison with the ending-guessing rules, the morphological rules
have much better precision and hence better accuracy of guessing.
Virtually almost every word which can be guessed by the morphological
rules is guessed exactly correct (97\% recall and 97\% precision).  Not
surprisingly, the coverage of morphological rules is much lower than that
of the ending-guessing ones -- for the suffix rules it is less than 40\%
and for the prefix rules about 5-6\%.

\begin{table*}[t]  

\begin{minipage}{\hsize}

\begin{tabular}[t]{||l|lll|lll||}
\hline\hline
Guessing   & \multicolumn{3}{c|}{Lexicon}    &  \multicolumn{3}{c|}{Corpus}   \\
Strategy   & Precision & Recall   & Coverage & Precision & Recall   & Coverage\\
\hline
Xerox (X)  & 0.460761  & 0.956331 & 0.917698 & 0.523965  & 0.944526 & 0.893275\\
Ending 75 
(E$_{75}$) & {\bf 0.675122}  
                       & 0.945726 & 0.977089 & {\bf 0.745339}  
                                                         & 0.952016 & 0.96104 \\
X+E$_{75}$ & 0.470249  & 0.95783  & 0.989843 & 0.519715  & 0.949789 & 0.969023 \\
E$_{75}$+X & 0.670741  & 0.943319 & 0.989843 & 0.743932  & 0.951541 & 0.969023 \\
\hline
P$_{80}$+
E$_{75}$   & 0.687126  & 0.946208 & 0.977488 & 0.748922  & 0.951563 & 0.96104 \\
S$_{60}$+
E$_{75}$   & 0.734143  & 0.945015 & 0.979686 & 0.792901  & 0.951015 & 0.963289 \\
P$_{80}$+S$_{60}$+
E$_{75}$   & {\bf 0.745504}  
                       & 0.945445 & 0.980086 & {\bf 0.796252}  
                                                         & 0.950562 & 0.963289 \\
\hline\hline
\end{tabular}

\vspace{2ex} 

\label{tab:eval_all}

\caption{Results of the cascading application of the rule-sets over the
  training lexicon and training corpus. P$_{80}$ - prefix rule-set scored
  over 80 points, S$_{60}$ - suffix rule-set scored over 60 points,
  E$_{75}$ - ending-guessing rule-set scored over 75 points.  As the
  base-line model was taken the ending guesser developed by Xerox (X).
  The first part of the table shows that the E$_{75}$ rule-set
  outperforms and fully supercedes the Xerox rule-set. The second part of
  the table shows that the cascading application of the morphological
  rule-sets together with the ending-guessing rules increases the
  performance by about 5\% in precision.  }

\longline

\end{minipage}
\end{table*}

After obtaining the optimal rule-sets we performed the same experiment on
a word-sample which was not included into the training lexicon and
corpus.  We gathered about three thousand words from the lexicon
developed for the Wall Street Journal corpus\footnote{these words were
  not listed in the training lexicon} and collected frequencies of these
words in this corpus.  At this experiment we obtained similar metrics
apart from the coverage which dropped about 0.5\% for Ending 75 and Xerox
rule-sets and 7\% for the Suffix 60 rule-set.  This, actually, did not
come as a surprise, since many main forms required by the suffix rules
were missing in the lexicon.

In the next experiment we evaluated whether the morphological rules add
any improvement if they are used in conjunction with the ending-guessing
rules. We also evaluated in detail whether a conjunctive application with
the Xerox guesser would boost the performance.  As in the previous
experiment we measured the precision, recall and coverage both on the
lexicon and on the corpus.  Table~2\ignore{ \ref{tab:eval_all} }
demonstrates some results of this experiment.  The first part of the
table shows that when the Xerox guesser is applied before the E$_{75}$
guesser we measure a drop in the performance. When the Xerox guesser is
applied after the E$_{75}$ guesser no sufficient changes to the
performance are noticed. This actually proves that the E$_{75}$ rule-set
fully supercedes the Xerox rule-set.  The second part of the table shows
that the cascading application of the morphological rule-sets together
with the ending-guessing rules increases the overall precision of the
guessing by a further 5\%.  This makes the improvements against the
base-line Xerox guesser 28\% in precision and 7\% in coverage.

\section{Tagging Unknown Words}

The direct evaluation of the rule-sets gave us the grounds for the
comparison and selection of the best performing guessing rule-sets.  The
task of unknown word guessing is, however, a subtask of the overall
part-of-speech tagging process. Thus we are mostly interested in how the
advantage of one rule-set over another will affect the tagging
performance.  So, we performed an independent evaluation of the impact of
the word guessers on tagging accuracy. In this evaluation we tried two
different taggers.  First, we used a tagger which was a {\sc c}++
re-implementation of the {\sc lisp} implemented HMM Xerox tagger
described in \cite{Kupiec:1992}.  The other tagger was the rule-based
tagger of Brill \cite{Brill:1995}.  Both of the taggers come with data
and word-guessing components pre-trained on the Brown
Corpus\footnote{Since Brill's tagger was trained on the Penn tag-set
  \cite{Marcus:1993} we provided an additional mapping.}.  This, actually
gave us the search-space of four combinations: the Xerox tagger equipped
with the original Xerox guesser, Brill's tagger with its original
guesser, the Xerox tagger with our cascading P$_{80}$+S$_{60}$+E$_{75}$
guesser and Brill's tagger with the cascading guesser.  For words which
failed to be guessed by the guessing rules we applied the standard method
of classifying them as common nouns (NN) if they are not capitalised
inside a sentence and proper nouns (NP) otherwise.  As the base-line
result we measured the performance of the taggers with all known words on
the same word sample.

In the evaluation of tagging accuracy on unknown words we pay attention
to two metrics.  First we measure the accuracy of tagging solely on
unknown words:

$Unkown Score = \frac{Correctly Tagged Unkown Words}{Total Unknown Words}$

This metric gives us the exact measure of how the tagger has done on
unknown words.  In this case, however, we do not account for the
known words which were mis-tagged because of the guessers. To put a perspective
on that aspect we measure the overall tagging performance:

$TotalScore = \frac{Correctly Tagged Words}{Total Words}$

\begin{table*}[t]  
\begin{minipage}{\hsize}
\begin{tabular}[t]{||l|l||l|l|l|l||l|l||}
\hline\hline
Tagger & Guessing & Total & Unkn. & Total  & Unkn.  &  Total  & Unkn. \\
       & strategy & words & words & mistag.& mistag.&  Score  & Score \\
\hline\hline
Xerox & Xerox    & 5,970  & 347   & 324    & 63     & 94.3\%  & 81.8\% \\
\hline
 Xerox & P$_{80}$
        +S$_{60}$
        +E$_{75}$& 5,970 & 347    & 292    & 33     & 95.1\%  & 90.5\% \\
\hline
Brill &  Brill   & 5,970 & 347    & 246    & 54     & 95.9\%  & 84.5\% \\
\hline
 Brill & P$_{80}$
        +S$_{60}$
        +E$_{75}$& 5,970 & 347    & 219    & 27     & 96.3\%  & 92.2\% \\
\hline\hline
\end{tabular}

\vspace{2ex} 

\label{tab:tag-1}

\caption{This table shows the results of tagging a  text with
  347 unknown words by four different combinations of two taggers and
  three word-guessing modules using the Brown Corpus model. The accuracy
  of tagging the unknown words when they were made known to the lexicon
  was detected at 98.5\% for both taggers.}

\longline
\end{minipage}
\end{table*}

\begin{table*}[t]  
\begin{minipage}{\hsize}
\begin{tabular}[t]{||l|l||l|l|l|l||l|l||}
\hline\hline
Tagger & Guessing & Total & Unkn. & Total   & Unkn.   &  Total & Unkn. \\
       & strategy & words & words & mistag. & mistag. &  Score & Score \\
\hline\hline
 Xerox & Xerox    & 5,970 & 2215  & 556     &  516    & 90.7\% & 76.7\% \\
\hline
 Xerox & P$_{80}$
        +S$_{60}$
        +E$_{75}$ & 5,970 & 2215  & 332     & 309     & 94.44\% &  86.05\%\\
\hline
Brill &  Brill    & 5,970 & 2215  & 464     & 410     & 93.1\%  &  81.5\%  \\
\hline
Brill & P$_{80}$
        +S$_{60}$
        +E$_{75}$ & 5,970 & 2215 & 327      & 287     & 94.52\% & 87.45\% \\
\hline\hline
\end{tabular}

\vspace{2ex} 

\label{tab:tag-2}

\caption{This table shows the results of tagging the same as in Table~3  text  by
  four different combinations of two taggers and three word-guessing
  modules using the Brown Corpus model and the lexicon which contained
  only closed-class and short words. The accuracy of tagging the unknown
  words when they were made known to the lexicon was detected at 90.3\%
  for the Xerox tagger and at 91.5\% for Brill's tagger.}

\longline
\end{minipage}
\end{table*}

Since the Brown Corpus model is a general language model, it, in
principle, does not put restrictions on the type of text it can be used
for, although its performance might be slightly lower than that of a
model specialised for this particular sublanguage.  Here we want to
stress that our primary task was not to evaluate the taggers themselves
but rather their performance with the word-guessing modules. So we did
not worry too much about tuning the taggers for the texts and used the
Brown Corpus model instead. We tagged several texts of different origins,
except from the Brown Corpus. These texts were not seen at the training
phase which means that neither the taggers nor the guessers had been
trained on these texts and they naturally had words unknown to the
lexicon. For each text we performed two tagging experiments. In the first
experiment we tagged the text with the Brown Corpus lexicon supplied with
the taggers and hence had only those unknown words which naturally occur
in this text. In the second experiment we tagged the same text with the
lexicon which contained only closed-class\footnote{articles,
  prepositions, conjunctions, etc.} and short\footnote{shorter than 5
  characters} words.  This small lexicon contained only 5,456 entries out
of 53,015 entries of the original Brown Corpus lexicon.  All other words
were considered as unknown and had to be guessed by the guessers.

We obtained quite stable results in these experiments.  Here is a typical
example of tagging a text of 5970 words.  This text was detected to have
347 unknown words. First, we tagged the text by the four different
combinations of the taggers with the word-guessers using the full-fledged
lexicon. The results of this tagging are summarised in
Table~3\ignore{\ref{tab:tag-1}}.  When using the Xerox tagger with its
original guesser, 63 unknown words were incorrectly tagged and the
accuracy on the unknown words was measured at 81.8\%. When the Xerox
tagger was equipped with our cascading guesser its accuracy on unknown
words increased by almost 9\% upto 90.5\%.  The same situation was
detected with Brill's tagger which in general was slightly more accurate
than the Xerox one\footnote{This, however, was not an entirely fair
  comparison because of the differences in the tag-sets in use by the
  taggers. The Xerox tagger was trained on the original Brown Corpus
  tag-set which makes more distinctions between categories than the Penn
  Brown Corpus tag-set.}.  The cascading guesser performed better than
Brill's original guesser by about 8\% boosting the performance on the
unknown words from 84.5\%\footnote{This figure agrees with the 85\%
  quoted by Brill \cite{Brill:1994}.} to 92.2\%.  The accuracy of the
taggers on the set of 347 unknown words when they were made known to the
lexicon was detected at 98.5\% for both taggers.

In the second experiment we tagged the same text in the same way but with
the small lexicon.  Out of 5,970 words of the text, 2,215 were unknown to
the small lexicon.  The results of this tagging are summarised in
Table~4\ignore{\ref{tab:tag-2}}. The accuracy of the taggers on the 2,215
unknown words when they were made known to the lexicon was much lower
than in the previous experiment --- 90.3\% for the Xerox tagger and
91.5\% for Brill's tagger. Naturally, the performance of the guessers was
also lower than in the previous experiment plus the fact that many
``semi-closed'' class adverbs like ``however'', ``instead'', etc., were
missing in the small lexicon.  The accuracy of the tagging on unknown
words dropped by about 5\% in general.  The best results on unknown words
were again obtained on the cascading guesser (86\%--87.45\%) and Brill's
tagger again did better then the Xerox one by 1.5\%.

Two types of mis-taggings caused by the guessers occured.  The first type
is when guessers provided broader \pos -classes for unknown words and the
tagger had difficulties with the disambiguation of such broader classes.
This is especially the case with the ``ing'' words which, in general, can
act as nouns, adjectives and gerunds and only direct lexicalization can
restrict the search space, as in the case with the word ``going'' which
cannot be an adjective but only a noun and a gerund.  The second type of
mis-tagging was caused by wrong assignments of \pos s by the guesser.
Usually this is the case with irregular words like, for example,
``cattle'' which was wrongly guessed as a singular noun (NN) but in fact
is a plural noun (NNS).

\section{Discussion and Conclusion}

We presented a technique for fully unsupervised statistical acquisition
of rules which guess possible parts-of-speech for words unknown to the
lexicon. This technique does not require specially prepared training data
and uses for training the lexicon and word frequencies collected from a
raw corpus. Using these training data three types of guessing rules are
learned: prefix morphological rules, suffix morphological rules and
ending-guessing rules.  To select best performing guessing rule-sets we
suggested an evaluation methodology, which is solely dedicated to the
performance of part-of-speech guessers.

Evaluation of tagging accuracy on unknown words using texts unseen by the
guessers and the taggers at the training phase showed that tagging with
the automatically induced cascading guesser was consistently more
accurate than previously quoted results known to the author (85\%). The
cascading guesser outperformed the guesser supplied with the Xerox tagger
by about 8-9\% and the guesser supplied with Brill's tagger by about
6-7\%.  Tagging accuracy on unknown words using the cascading guesser was
detected at 90-92\% when tagging with the full-fledged lexicon and
86--88\% when tagging with the closed-class and short word lexicon.  When
the unknown words were made known to the lexicon the accuracy of tagging
was detected at 96-98\% and 90-92\% respectively.  This makes the
accuracy drop caused by the cascading guesser to be less than 6\% in
general.  Another important conclusion from the evaluation experiments is
that the morphological guessing rules do improve the guessing
performance.  Since they are more accurate than ending-guessing rules
they are applied before ending-guessing rules and improve the precision
of the guessings by about 5\%. This, actually, results in about 2\%
higher accuracy of tagging on unknown words.

The acquired guessing rules employed in our cascading guesser are, in
fact, of a standard nature and in that form or another are used in other
\pos -guessers.  There are, however, a few points which make the
rule-sets acquired by the presented here technique more accurate:
\begin{itemize}
\item the learning of such rules is done from the lexicon rather than
  tagged corpus, because the guesser's task is akin to the lexicon
  lookup;
\item there is a well-tuned statistical scoring procedure which accounts
  for rule features and frequency distribution;
\item there is an empirical way to determine an optimum collection of
  rules, since acquired rules are subject to rigorous direct evaluation
  in terms of precision, recall and coverage;
\item rules are applied cascadingly using the most accurate rules first.
\end{itemize}

One of the most important issues in the induction of guessing rule-sets
is the choice right data for training.  In our approach, guessing rules
are extracted from the lexicon and the actual corpus frequencies of
word-usage then allow for discrimination between rules which are no
longer productive (but have left their imprint on the basic lexicon) and
rules that are productive in real-life texts.  Thus the major factor in
the learning process is the lexicon. Since guessing rules are meant to
capture general language regularities the lexicon should be as general as
possible (list {\em all} possible \pos s for a word) and as large as
possible.  The corresponding corpus should include most of the words from
the lexicon and be large enough to obtain reliable estimates of
word-frequency distribution.  Our experiments with the lexicon and word
frequencies derived from the Brown Corpus, which can be considered as a
general model of English, resulted in guessing rule-sets which proved to
be domain and corpus independent\footnote{but tag-set dependent},
producing similar results on test texts of different origin.

Although in general the performance of the cascading guesser is only 6\%
worse than the lookup of a general language lexicon there is room for
improvement.  First, in the extraction of the morphological rules we did
not attempt to model non-concatenative cases. In English, however, since
most of letter mutations occur in the last letter of the main word it is
possible to account for it. So our next goal is to extract morphological
rules with one letter mutations at the end.  This would account for cases
like ``try - tries'', ``reduce - reducing'', ``advise - advisable''.  We
expect it to increase the coverage of thesuffix morphological rules and
hence contribute to the overall guessing accuracy.  Another avenue for
improvement is to provide the guessing rules with the probabilities of
emission of \pos s from their resulting \pos -classes. This information
can be compiled automatically and also might improve the accuracy of
tagging unknown words.

The described rule acquisition and evaluation methods are implemented as
a modular set of {\sc c}++ and {\sc awk} tools, and the guesser is easily
extendable to sub-language specific regularities and retrainable to new
tag-sets and other languages, provided that these languages have
affixational morphology.  Both the software and the produced guessing
rule-sets are available by contacting the author.

\section{Acknowledgements}
Some of the research reported here was funded as part of {\sc epsrc}
project IED4/1/5808 ``Integrated Language Database''.  I would also like
to thank Chris Brew for helpful discussions on the issues related to this
paper.

\end{document}